\documentclass[letterpaper, 10 pt, conference]{ieeeconf}  % Comment this line out if you need a4paper

\IEEEoverridecommandlockouts                              % This command is only needed if 
\usepackage{cite}
\usepackage{amsmath,amssymb,amsfonts}
\usepackage{dsfont}
\usepackage{algorithmic}
\usepackage{graphicx}

\usepackage{caption}
\usepackage{subcaption}
\usepackage{textcomp}
\usepackage{xspace}
\usepackage{balance}
\usepackage{paralist}
\usepackage{url}
\usepackage[bottom]{footmisc}
\usepackage{hyperref}
\usepackage[dvipsnames]{xcolor}
\usepackage{tabularx} % for 'tabularx' environment
\usepackage{multirow}

\title{\LARGE \bf
MTBF Model for AVs - From Perception Errors to Vehicle-Level Failures
}

\author{Fabian Oboril$^{1}$, Cornelius Buerkle$^{1}$, Alon Sussmann$^{2}$, Simcha Bitton$^{2}$, Simone Fabris$^{2}$
\thanks{$^{1}$Intel Corporation, Intel Labs. 
        {\tt\small \{cornelius.buerkle, fabian.oboril\}@intel.com}}%
\thanks{$^{2}$Mobileye.
        {\tt\small \{alon.sussmann,simcha.c, simone.fabris\}@mobileye.com}}%
}

\begin{document}
\bstctlcite{IEEEexample:BSTcontrol}

\maketitle
\thispagestyle{empty}
\pagestyle{empty}

%%%%%%%%%%%%%%%%%%%%%%%%%%%%%%%%%%%%%%%%%%%%%%%%%%%%%%%%%%%%%%%%%%%%%%%%%%%%%%%%
\begin{abstract}
The development of Automated Vehicles (AVs) is progressing quickly and the first robotaxi services are being deployed worldwide. However, to receive authority certification for mass deployment, manufactures need to justify that their AVs operate safer than human drivers. This in turn creates the need to estimate and model the collision rate (failure rate) of an AV taking all possible errors and driving situations into account. In other words, there is the strong demand for comprehensive Mean Time Between Failure (MTBF) models for AVs. In this paper, we will introduce such a generic and scalable model that creates a link between errors in the perception system to vehicle-level failures (collisions). Using this model, we are able to derive requirements for the perception quality based on the desired vehicle-level MTBF or vice versa to obtain an MTBF value given a certain mission profile and perception quality.
\end{abstract}

%%%%%%%%%%%%%%%%%%%%%%%%%%%%%%%%%%%%%%%%%%%%%%%%%%%%%%%%%%%%%%%%%%%%%%%%%%%%%%%%
\section{Introduction}

Automated Vehicles (AVs) are the greatest revolution in the transportation industry since the introduction of the internal combustion engine. Consequently, many hopes are linked to the success of AVs, for example improved road safety~\cite{shalev2018vision}. %However, also various challenges have to be addressed to make mass deployments of AVs reality. 
Thanks to great progress over the past years, the first use of AVs beyond pure research prototypes is finally possible. For example, Mobileye announced recently their first robotaxi solution to hit the market in 2022~\cite{robotaxi2022}.

Yet, not all challenges are resolved yet, and a major one is still demanding attention: safety assurance and validation~\cite{SaFAD2019}. In fact, it is required to prove that an AV is safe w.r.t the requirements specified in ISO 26262 (\textit{functional safety} - FuSa) and \textit{safety of the intended functionality} (SOTIF, ISO 21448), to receive authority certification for public use. %While FuSa addresses safety w.r.t failures of a (sub)system, SOTIF considers hazards due to inherent limitations or weaknesses of a (sub)system such as limitations of the perception to detect an object properly. % As a result, SOTIF requires estimation of the \textit{residual risk} of the entire system considering possible trigger conditions and system weaknesses~\cite{ansysSOTIF} and adequate safety goals need to be defined.

Consequently, adequate safety goals need to be defined throughout the development cycle and appropriate validation and verification mechanisms have to be installed. A very important safety goal to achieve trust in AV systems is that AVs drive ``safer'' than human drivers. In fact, for example the German Federal Ministry of Transport and Digital Infrastructure states~\cite{di2017ethics}: ``The licensing of
automated systems is not justifiable unless it promises to produce at least a diminution in harm compared with human driving [...]''. %
%
%In order that the public can trust an AV system, the AV requires a significant increase with respect to driving safety compared to human drivers, which means that the residual risk of harm due to collisions is acceptably low\footnote{German Federal Ministry of Transport and Digital Infrastructure states~\cite{di2017ethics}: ``The licensing of automated systems is not justifiable unless it promises to produce at least a diminution in harm compared with human driving [...]''.}. 
%For this reason, it is desired that an AV is involved in fewer (severe) accidents than a human driver. 
In this regard, a good target for AV performance is to be 10 to 100 times better than a human driver, who has on average every $10^5$ hours a severe accident~\cite{weastMTBF}. Hence, the corresponding AV safety goal is a \textit{Mean Time Between Failure (MTBF)} of $10^6$ or $10^7$ hours (or better), where vehicle-level failures are defined as collisions.

At the same time, a complex system such as an AV, that operates in a highly dynamic environment, can fail due to many reasons, for example planning errors can maneuver the AV into a dangerous situation, or perception errors can cause undetected objects. Fortunately, errors in planning components can be mitigated with safety frameworks such as ``Responsibility Sensitive Safety'' (RSS)~\cite{shalev2017formal,ieeeAVSafety}, which can ensure safe operation of the AV, if the perception is error-free. On the other hand, comprehensive safety solutions for the perception systems are still missing. To make matters worse, even the best perception systems have inherent weaknesses and limitations, which can cause undetected objects even in the close vicinity of the AV, as shown in~\cite{buerkle2021safe}. If such a perception error appears in a critical traffic situation (e.g. non-detected standing object in front of the AV), a hazardous event (i.e. collision) can be the consequence.

Hence, to ensure that the AV safety goals can be achieved in the field, it is necessary to comprehensively model the relation between the perception error rate and the resulting vehicle-level failure rate\footnote{as explained planning errors can be excluded if e.g. RSS is in place}, while considering masking factors due to other components or due to traffic situation. %For example, if the perception system failed to detect a vehicle, which moves away from the AV, it might not result in any consequences. 
%In other words, a solution is required to estimate residual risk at vehicle level due to perception errors. 
Unfortunately, such a solution is still missing.

\begin{figure}[t!]
\includegraphics[width=\columnwidth]{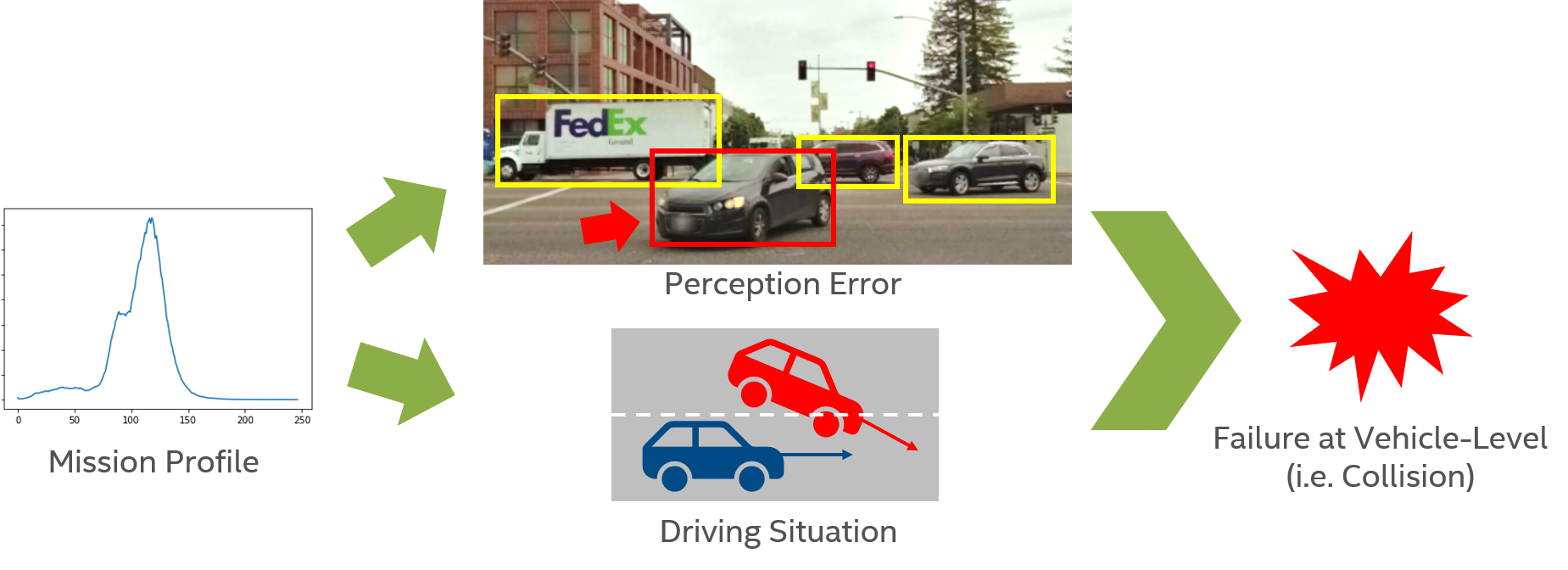}
%\vspace{-0.25cm}
\caption{Perception errors can only cause vehicle-level failures (i.e. collisions) in critical driving situations, and the rates of both depend on the vehicle mission profile (speed range, road characteristics, traffic density, time of day, etc.).} %Hence, we propose a generic approach to derive vehicle-level failure rates based considering both aspects.}
\label{fig:mtbf_idea}
\vspace{-0.5cm}
\end{figure}

For this reason, \textit{we propose in this paper a novel, generic approach to link vehicle-level failure rates to perception error rates}. This approach considers not only perception errors, but also the likelihoods of an AV being in a potentially relevant traffic situation and various masking effects given a certain mission profile\footnote{A mission profile is a simplified representation of all relevant static and dynamic conditions that a vehicle component is exposed to within its lifecycle, i.e. temperature, operational times, driving speeds, traffic conditions, road types, etc.} (see Figure~\ref{fig:mtbf_idea}). This enables us to derive a vehicle-level MTBF estimation given mission profiles and perception error rates, which can be obtained from manufacturer or public databases, as we will show. In addition, the approach also allows us to derive requirements for the perception quality based on a desired vehicle-level MTBF. For instance, our results based on naturalistic driving data collected on German highways show that a perception error rate below $10^{-5}$ is required to achieve the desired safety goal of an MTBF of $10^6$ hours.

The remainder of the paper is organized as follows. In Section~\ref{sec:prelim}, we introduce some preliminary assumptions on the AV operation. In Section~\ref{sec:approach}, the new approach for vehicle-level failure rate estimation is presented, followed by an exemplary study in Section~\ref{sec:results}. Finally, Section~V concludes the paper.

\section{Preliminaries}
\label{sec:prelim}

\begin{figure}[t!]
\centering
\vspace{-0.1cm}
\includegraphics[width=0.9\columnwidth]{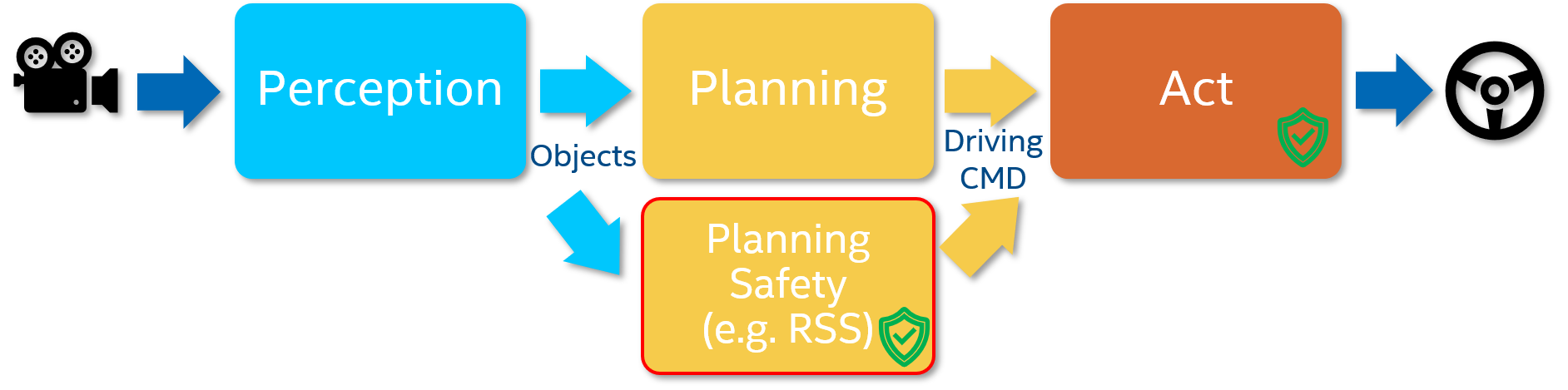}
\caption{Basic AV Pipeline}
\label{fig:av_setup}
\vspace{-0.5cm}
\end{figure}

Most of the AVs operate based on the same common principle: Perceive, Plan, Act. A variety of sensors is used by the ``Perception'' system to capture the environment and create a digital representation, the so-called environment model. This model is used by the ``Planning'' component to identify the next set of actions and behaviors, which are then executed by the ``Act'' subsystem, as depicted in Figure~\ref{fig:av_setup}.

In terms of safety assurance, the ``Act'' system typically relies on hardware redundancy to ensure proper execution of the planned driving commands. In addition, the safety approaches for these components are well established and manufacturers as well as certification authorities understand the necessities. For ``Perceive'' and ``Plan'' the situation is different. To ensure safety of an AV planning system, the \textit{Responsibility Sensitive Safety (RSS)} approach\footnote{RSS is a formal, physics-based model representing common sense rules for safe and responsible driving behavior.} was proposed a few years ago~\cite{shalev2017formal}, and has been contributed to the upcoming IEEE 2846 standard~\cite{ieeeAVSafety}.  %As these models are parametric, it is possible to choose reasonable parameter values in a mission-dependent manner to always maintain the required level of safety, or in other words to keep the failure rate of the system (i.e. number of collisions per driven hour) below an acceptable level.

However, industry must find similar and adequate (SOTIF) solutions on perception systems that allow to capture the influence of perception errors on vehicle-level failures and by that mean can be used to prove that the perception system is sufficiently safe. However, while lots of effort is spent on improving the overall perception quality, and thus reduce error rates, the inherent limitations of perception systems are not yet solved. Therefore, still false alarms (detection of non-existing objects) or detection misses (non-detection of an existing object) may occur. These errors then propagate through the rest of the pipeline and may result in a wrong or inadequate driving decision (command). Subsequently, the vehicle may be forced by the planning stage (incl. its safety companion) to perform an emergency stop in case of a perception false alarm (Plan assumes an unsafe state although safe), or not decelerate properly in case of a detection miss (Plan assumes a safe state although unsafe). 

However, despite its importance, the relation among perception errors and vehicle collisions has not been sufficiently studied, so far. In~\cite{weastMTBF} the need to estimate vehicle-level MTBF was discussed, yet a generic and scalable model is missing. The authors in~\cite{buerkle2021safe} introduced the notion of safety-relevant perception errors and underlined that not every perception error will lead to a collision. However, also in this work, a concrete model capturing the relation between collisions and perception errors is missing. Similarly, the work in~\cite{berk2019safety} explores probabilistic modeling approaches to quantify the perception reliability, yet also does not study if a perception error can cause a collision or not.

In this work, we intend to close this gap and present a generic and scalable model, that captures the impact of perception errors together with mission profile and the related traffic situations on the overall vehicle-level failure rate. 

%Please note that the approach which will be presented in the next section is generic and could be extended to also include errors in the planning domain.

%\begin{figure}[t!]
%\includegraphics[width=\columnwidth]{images/sotif_flow.png}
%\caption{bla}
%\label{fig:sotif}
%\end{figure}

%
%For this reason, we make the following assumptions on the AV system:
%\begin{itemize}
%\item The AV uses RSS or similar, and thus neither ``Plan'' nor ``Act'' can fail.
%\item A vehicle-level failure is a collision with high severity (see [todo])
%\item A perception error is critical if and only if it causes RSS (or similar) to judge the situation as safe, although it is unsafe, i.e. the vehicle would need to re-act on an object but will not do so due to the perception error.
%\item 
%\end{itemize}
%
%- Severity classification
%- SOTIF figure?

\section{Vehicle-Level Failure Rate Estimation}
\label{sec:approach}

In this section, we present our approach to link perception errors to vehicle-level failures, i.e. collisions. For this purpose, it is first necessary to understand which perception errors at all can cause collisions and under which environmental circumstances (i.e. traffic situations) this may happen (see Figure~\ref{fig:mtbf_idea}). Therefore, we start with a set of necessary definitions, that help us to derive the model in the third part of this section.

\subsection{Safety Relevant and Severe Perception Errors}

As explained in Section~\ref{sec:prelim}, perception errors can occur even in the best perception systems, due to inherent limitations, which in consequence can lead to an unsafe vehicle behavior. In this regard, it is important to differentiate the different flavors of perception errors that can occur: First, there are perception misses, also called \textit{False Negative} errors, where an object is not detected. The opposite case is possible as well, where the perception system reports a non-existing object (\textit{False Positive} error / False alarm). Second, the perception system can provide wrong distance or velocity values, that result either in an object that is in reality closer or slower than perceived, or that is further away or faster than reported. As the impact on vehicle-level safety of the first type of velocity or distance errors is similar to perception misses (in both cases the system perceives a situation as safe, which might be unsafe), we classify both as \textit{Type II errors} (rejection of an actually true null hypothesis). False Positive errors and the other velocity and distance errors belong to \textit{Type I errors} (acceptance of an actually false null hypothesis), as for these an actually safe situation, may be perceived as unsafe. As a result, Type I errors can cause for instance unwanted braking maneuvers, while Type II errors can lead to a vehicle not braking although it should. \vspace{+0.1cm}
\paragraph*{\textbf{Definition 1: Type I and Type II Perception Errors}}
~\newline A perception error is of:\begin{itemize}
\item \textit{Type I}, if the perceived object information is more severe than it is in reality (i.e. overestimating the risk). % (e.g. unsafe although safe).
\item \textit{Type II}, if the perceived object information is less severe than it is in reality (i.e. underestimating the risk). % (e.g. safe although unsafe).
\end{itemize}

It is important to note that not every perception error affects the safety judgment of the AV planning subsystem, for example a false alarm on a neighboring lane, not within the path of the AV, may not trigger a vehicle action. In fact, only a subset of errors are truly relevant; those that can change the safety judgment of the planning stage (e.g. not sufficient braking or unnecessary braking). Therefore, we use the following definition for \textit{safety-relevant perception errors}:\vspace{+0.1cm}
\paragraph*{\textbf{Definition 2: Safety-Relevant Perception Error}}
~\newline A perception error is safety-relevant if and only if it changes the safety decision of the AV planning module:\begin{itemize}
\item causing an unnecessary emergency maneuver (e.g. strong deceleration), or
\item resulting in the AV not performing the required emergency maneuver (e.g. insufficient braking)
\end{itemize}
Note that the first case can only happen for Type~I errors, while Type II errors are linked to the second case.

\begin{figure}[b!]
\vspace{-0.3cm}
\centering
\includegraphics[width=0.85\columnwidth]{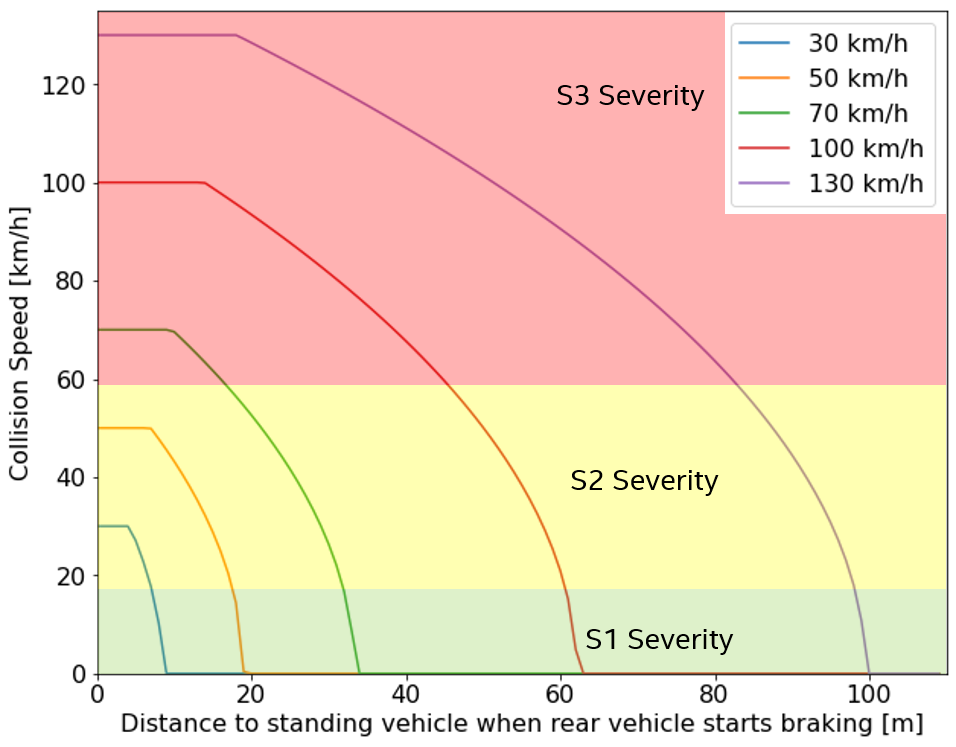}
\caption{Rear-end collision severity depending on vehicle speed and distance for a car-following situation (front vehicle stands still, rear vehicle decelerates with $8m/s^2$ after 0.5s reaction time)}
\label{fig:severity}
\vspace{-0.1cm}
\end{figure}

By changing the safety decision of the planning system, the result of a perception error on vehicle-level can be a collision with another object. As not every collision is equally severe, we further differentiate errors based on their potential result on vehicle-level. Some errors may just cause material damage, while others may result in severe harm to humans. In this regard, the ISO 26262 defines four classes of severity, S0 to S3 from no injuries upto fatal events (see an illustration of severity classes based on collision speed in Figure~\ref{fig:severity}). Thus, we further differentiate safety-relevant perception errors among severe and non-severe errors.
\vspace{+0.1cm}
\paragraph*{\textbf{Definition 3: Severe Perception Error}}
~\newline A perception error is severe if and only if it is a safety-relevant perception error that can cause a collision with S2 or S3 severity according to ISO 26262.

\subsection{Potentially Dangerous Traffic Situation}
Besides a safety-relevant perception error, it is also required that a \textit{potentially dangerous traffic situation} takes place to cause a collision (see Figure~\ref{fig:mtbf_idea}). For example, a perception miss of a leading vehicle may not have any consequences, if the leading vehicle is faster than the AV. For this reason, we use the following definition for potentially dangerous situations:\vspace{+0.1cm}
\paragraph*{\textbf{Definition 4: Potentially Dangerous Traffic Situation}}
A traffic situation is potentially dangerous if and only if a perception error (Type I or Type II) will cause a collision.
\vspace{+0.1cm}

\begin{figure}[b!]
\vspace{-0.2cm}
\includegraphics[width=\columnwidth]{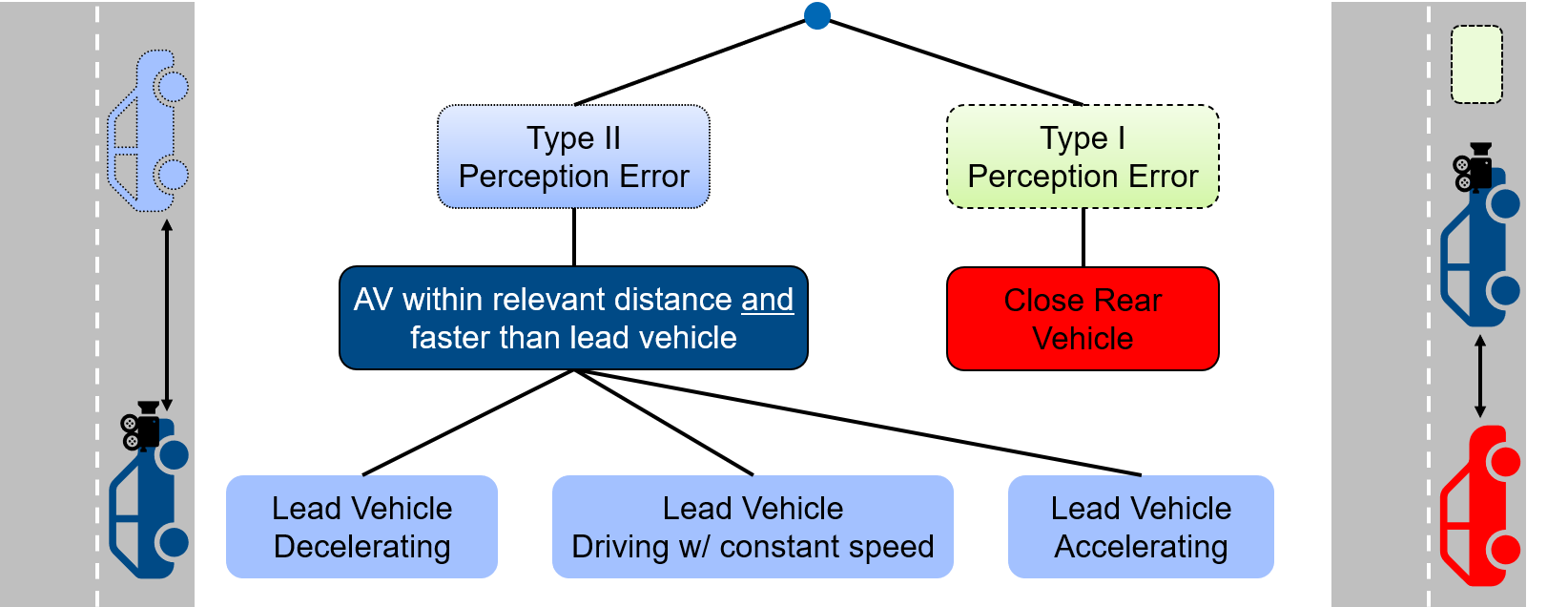}
\caption{Relation of potentially dangerous driving situations and Type I/II perception errors for a lane following scenario}
\label{fig:rds}
\vspace{-0.1cm}
\end{figure}

An intuitive example for a potentially dangerous traffic situation for a Type I perception error is a vehicle that follows the AV very closely, and thus cannot react to an unexpected braking maneuver of the AV caused by a false alarm of the AV's perception system. For Type~II errors, an easy example is an AV approaching an undetected standing vehicle, where the AV does not decelerate appropriately due to a detection miss. As shown by these examples, the type of traffic situation is correlated to the perception error class. For the case of a lane-following situation (e.g. on a highway), Figure~\ref{fig:rds} depicts this in more detail.

\subsection{Statistical Model}
As we have seen in the previous subsections, \textit{a collision (vehicle-level failure) is the combination of a perception error and a potentially dangerous traffic situation} (see Figure~\ref{fig:mtbf_idea}). Both events can be considered as independent, which means that the vehicle-level failure rate $\lambda$ can be estimated by multiplying the perception error rate $\lambda_p$ and the probability of being within a relevant traffic situation $p_S$. However, as there are two different types of perception errors, the overall failure rate is the sum of the products, as given in Equation~\eqref{eq:simpleModel}.
\begin{equation}
\lambda = \sum_{t\in\left\lbrace1,2\right\rbrace}  \lambda_{p_t} \times p_{S_t}
\label{eq:simpleModel}
\end{equation}

Having the overall failure rate $\lambda$ at vehicle-level, the \textit{Mean Time Between Failure} (MTBF) for the AV is defined as the inverse, i.e:
\begin{equation}
\text{MTBF} = \lambda^{-1}
\end{equation}

$\lambda_p$ and $p_S$ are dependent on the mission profile, for example if a vehicle is supposed to be deployed on highways or mainly in urban conditions. Moreover, also the expected speed ranges play an important role as we will see in Section~\ref{sec:results}. Hence, we propose to extend the simple model of Equation~\eqref{eq:simpleModel} to a model that captures $n$ speed ranges and different mission profiles (other extensions are also possible in a similar way). For a mission profile $m$ and a speed range $i\in\left\lbrace r_1,\dots,r_n\right\rbrace$, the extended model is:
\begin{equation}
\begin{split}
\lambda &= \sum_m \lambda_m \\
        &= \sum_m p_m \sum_i p_{i,m} \left[ \sum_{t\in\left\lbrace1,2\right\rbrace}  \lambda_{p_{t,m,i}} \times p_{S_{t,m,i}} \right]
\end{split}
\label{eq:extendedModel}
\end{equation}
Here, $p_m$ is the occurrence probability of mission profile $m$, and similarly $p_{i,m}$ is the occurrence probability of the speed range $i$ (e.g. the range from 100\,km/h to 130\,km/h) for this mission profile.

\begin{figure}[t!]
\includegraphics[width=\columnwidth]{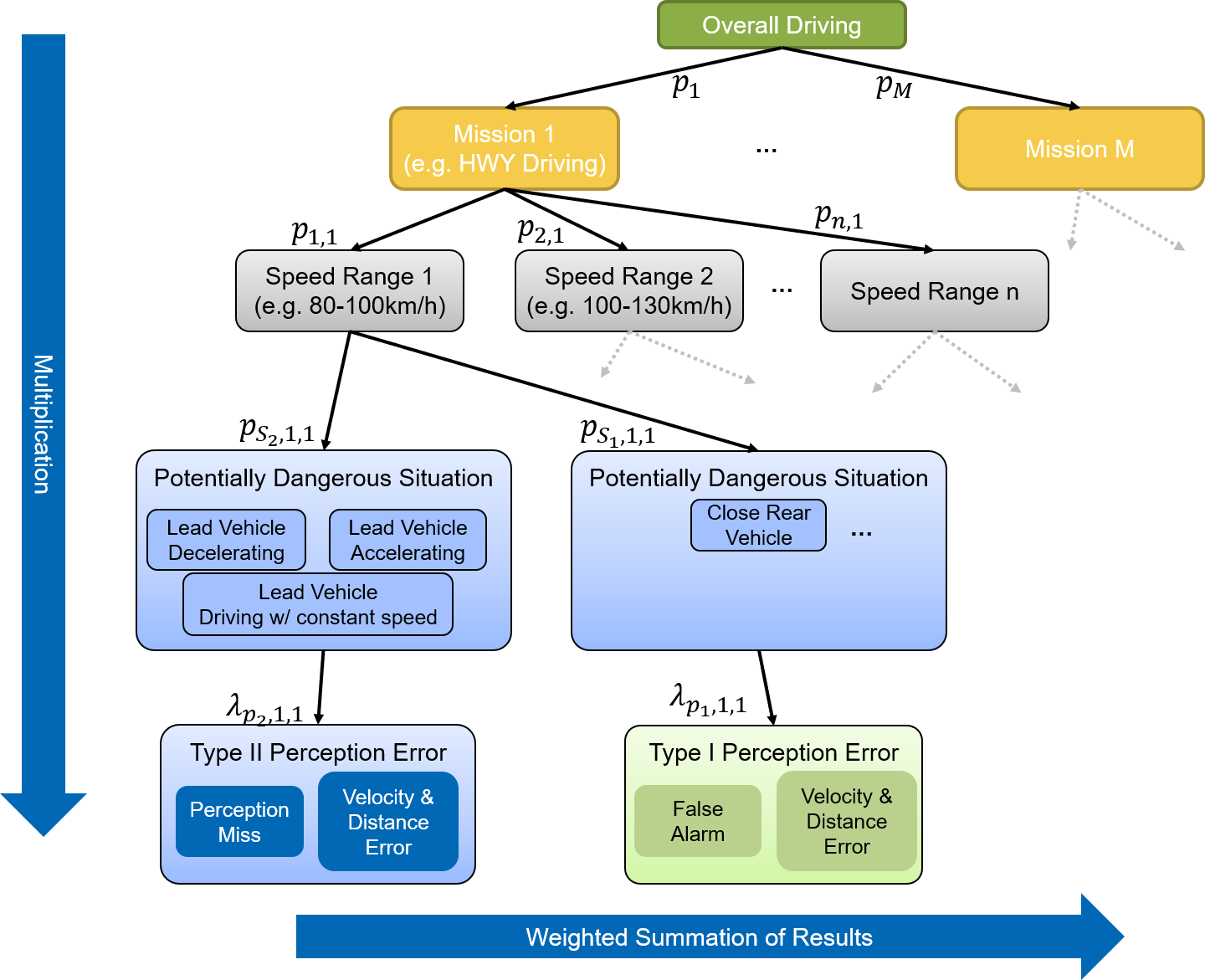}
\caption{Illustration of our proposed statistical model as probability tree to obtain the overall failure rate at vehicle-level linked to perception errors}
\label{fig:model}
\vspace{-0.5cm}
\end{figure}

\begin{figure}[t!]
\centering
\includegraphics[scale=0.4]{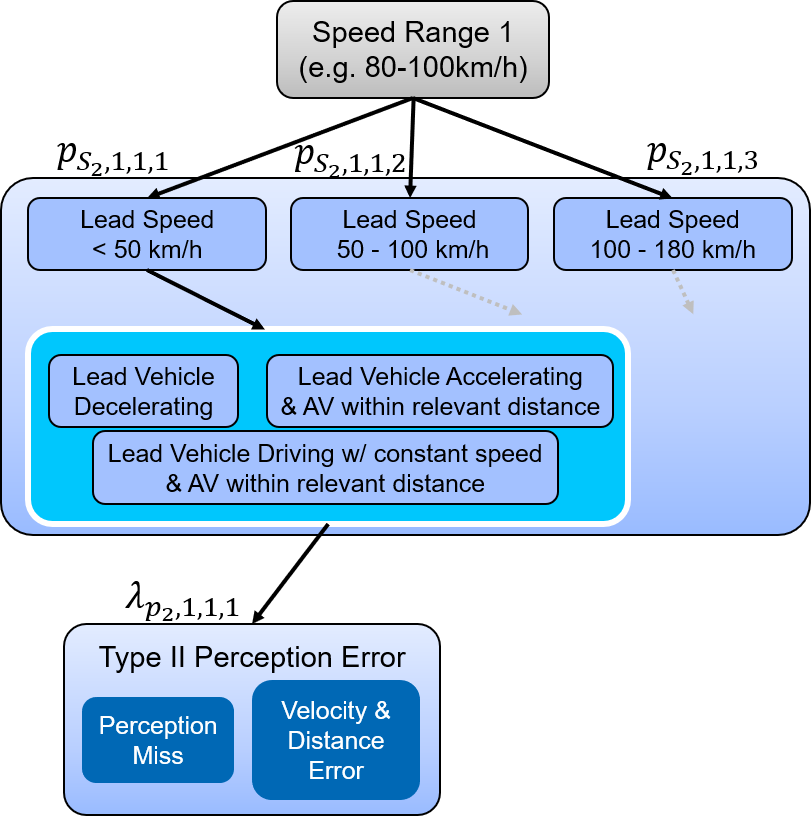}
\caption{Model refinement to separate potentially dangerous traffic situations for Type II perception errors based on the speed of the leading vehicle}
\label{fig:model_refinement}
\vspace{-0.5cm}
\end{figure}

Figure~\ref{fig:model} shows an illustration of our envisioned model as probability tree, where the edges represent the conditional probabilities that should be multiplied in vertical direction and added horizontally. In this regard, it is important to mention that our model is scalable and generic, i.e. it does not come with any restrictions on how mission profiles or speed ranges should be clustered. Based on the results presented in Section~\ref{sec:results}, we recommend to use different speed ranges, but if the overall amount of data is limited there might be situations where just one overall speed bucket or mission profile is the better choice. Furthermore, the potentially dangerous situations can also be further subdivided, so can it be done for the perception errors, as well. For instance, the duration of false alarms might be considered (with adequate sub-trees), or the behavior of the other vehicle can be further classified depending on its speed as illustrated in Figure~\ref{fig:model_refinement}. The reason for this refinement is that the speed of the leading vehicle is an important influencing factor, when evaluating the severity of a Type II perception error: The faster the AV and the slower the leading vehicle, the more likely a Type~II perception error will become severe. %Moreover, we also make use of a restriction for the potentially dangerous situations, that the AV and the other vehicle shall be within a relevant distance (e.g. 5\,s time-to-collision), to avoid consideration of irrelevant situations.

Please note that the model introduced above, can also be explained mathematically. For this, the number of perception errors $X_i$ is modeled as Poisson distribution
\begin{equation}
X_i \sim \text{Pois}(\lambda_{p}).
\end{equation}

Modeling the driving situations as Bernoulli distribution, it follows that the number of failures can be formulated as
\begin{equation}
F_i = \sum_{k=1}^{X_i} Z_k = \sum_{k=1}^{\infty} Z_k \times \mathds{1}_{k\leq X_i},
\end{equation}
where $Z_k \sim \text{ber}(p_S)$, and $p_S$ is the accumulated probability of being in a relevant driving situation. As $X_i$ and $Z_k$ can be assumed to be independent for every $k$, it follows that the expected value of $F_i$ is
\begin{equation}
\begin{split}
E(F_i) &= \sum_{k=1}^{\infty} E(Z_k) \times E(\mathds{1}_{k\leq X_i}) \\
       &= p_S \times E(X_i) = p_S \times \lambda_{X_i} = p_S \times \lambda_{p}  = \lambda
\end{split}
\end{equation}
This is where we started in Equation~\eqref{eq:simpleModel}.

Finally, it is worth noting that the perception error rate $\lambda_p$ is the superposition of errors due to hardware faults (e.g. a perception miss due to a radiation-induced bit flip) and errors originating from software (e.g. perception miss due to a not-well trained AI model). If it is desired to separate both a specific hardware failure rate and software failure rate can be considered by our model.

\subsection{Model Input}
\begin{figure}[t!]
\centering
\includegraphics[scale=0.4]{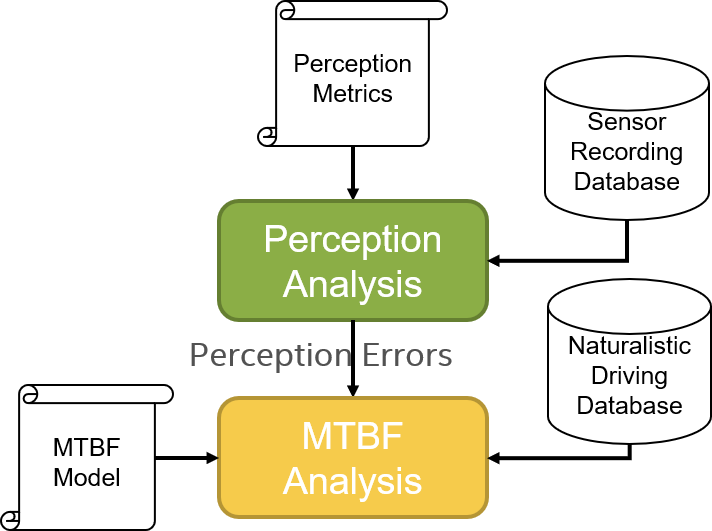}
\caption{Analysis flow: First obtain perception error (rates) from a comprehensive analysis on sensor recordings; Then estimate MTBF with naturalistic driving data to model potentially dangerous situations}
\label{fig:mtbf_approach}
\vspace{-0.55cm}
\end{figure}

An important aspect for our envisioned vehicle-level failure rate estimation model is the source for the input data. As the model itself is generic, it can be fed with data gathered from data recordings, simulation, public datasets or even on-the-fly while driving. However, to provide meaningful results, we envision to use comprehensive datasets and recordings as input, as depicted in Figure~\ref{fig:mtbf_approach}.

The occurrence probabilities of potentially dangerous situations can be for instance obtained by analyzing naturalistic driving datasets such as HighD~\cite{krajewski2018highd}, or by using already gathered statistical data e.g. from the NHTSA. For example the authors of~\cite{li2015driving} used naturalistic driving data for highways and found that the probability of being in a situation relevant for Type II perception errors is around 45\,\%. Similarly, we used the HighD dataset for our experimental study presented in Section~\ref{sec:results}. %In this regard, it is worth mentioning that a more comprehensive dataset allows a more accurate analysis, and 

To obtain perception error rates, the Definitions 1 to 4 need to be converted to quantifiable performance indicators, which are then obtained by testing the perception system against simulation, data recordings or directly on the road. To illustrate how this can be done, let us use Type II perception errors as example (for Type I errors it is similar). As specified in Definition 2, a relevant perception error will flip the safety consideration of the AV planning system. Assuming that this system acts according to RSS~\cite{shalev2017formal}, the number of relevant perception Type II errors for car-following situations can be obtained by
\begin{equation}
\sum_e \mathds{1}_{d_{per} > d_{RSS} > d_{real}},
\end{equation}
where $e$ are all evaluated events (e.g. clips, frames, etc.), and $\mathds{1}_{d_{per} > d_{RSS} > d_{real}}$ is 1 if $d_{per} > d_{RSS} > d_{real}$, else 0. In this regard, $d_{per}$ is the perceived distance between the AV and the leading vehicle, $d_{RSS}$ is the required safety distance according to RSS, and $d_{real}$ is the distance in reality. In other words, the perception error makes the system to judge the system as safe ($d_{per} > d_{RSS}$), while in reality it is unsafe ($d_{real} < d_{RSS}$). In case an object is not detected at all, one may assume that $d_{per} = \infty$.

%, we suggest to use a worst-case assumption on its velocity when calculating the necessary safety distance (according to RSS). In case of a car-following situation this is that the object stands still.

\section{Experimental Results}
\label{sec:results}

In order to show how our proposed model can be used, and also show its practical results, we applied it to the use case (mission profile) of highway driving (i.e. speed range between 80\,km/h and 180\,km/h), restricted to lane following situations. We used the HighD dataset~\cite{krajewski2018highd} to obtain naturalistic driving data for highways (in Germany), which forms the basis to derive the occurrence probabilities $p_S$ for the potentially dangerous situations. %To get a realistic assumption on the perception error rates, we used a LiDAR based object detector, called PointPillars~\cite{PointPillars}, which we evaluated on the Lyft perception dataset~\cite{LyftDS} in default of a highway perception dataset. 

\begin{figure}[t!]
\centering
\includegraphics[width=0.8\columnwidth]{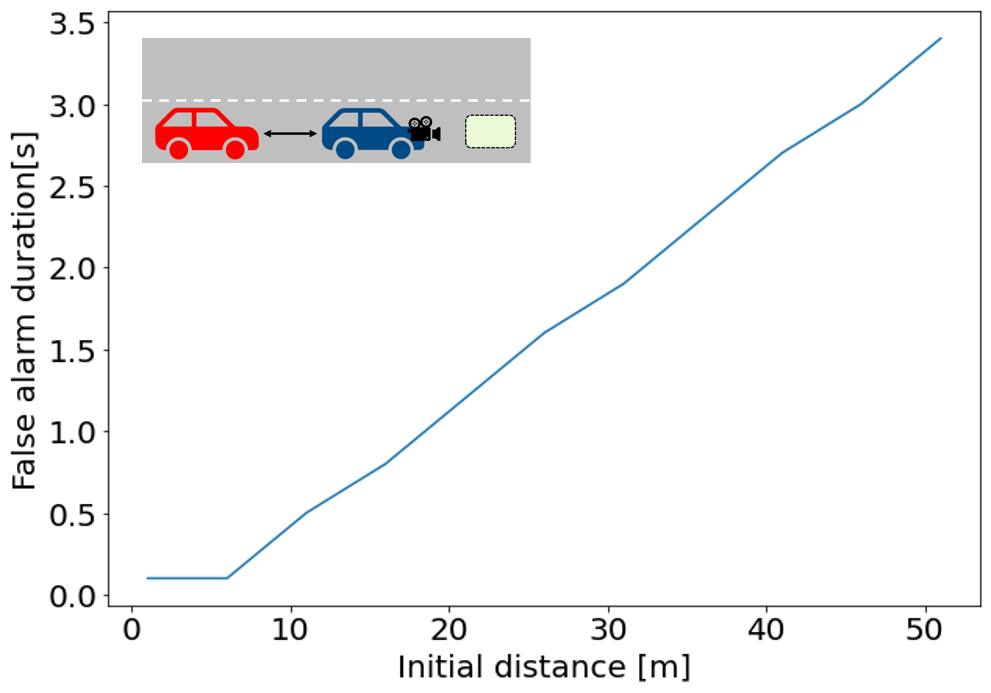}
\caption{Duration of a false alarm required to cause a S2 or S3 collision in case the false alarm causes the lead vehicle to perform a braking maneuver with $8m/s^2$ (lead and rear vehicle drive with 130km/h)}
\label{fig:fp_duration}
\vspace{-0.55cm}
\end{figure}

An important aspect that requires attention is that we will focus on Type II perception errors in this analysis. The reason is that the traffic constellations in the HighD dataset are such that Type I errors, for example false alarms, have to be present for more than 1 second, and have to cause a persistent emergency brake maneuver of the AV to cause noticeable consequences (see Figure~\ref{fig:fp_duration}). For example, a persistent false alarm of 1 second can only cause a S2 or S3 collision, if the rear vehicle follows with less than 20\,m distance at 130\,km/h. It follows that the impact of Type I errors on vehicle-level safety plays an inferior role compared to Type II errors for the mission profile used in this study. Similarly, velocity and distance errors are negligible compared to perception misses, thus we focus in this section only on the latter.

\begin{figure*}[t!]
\begin{subfigure}{.32\textwidth}
  \centering
  % include first image
  \includegraphics[width=\columnwidth]{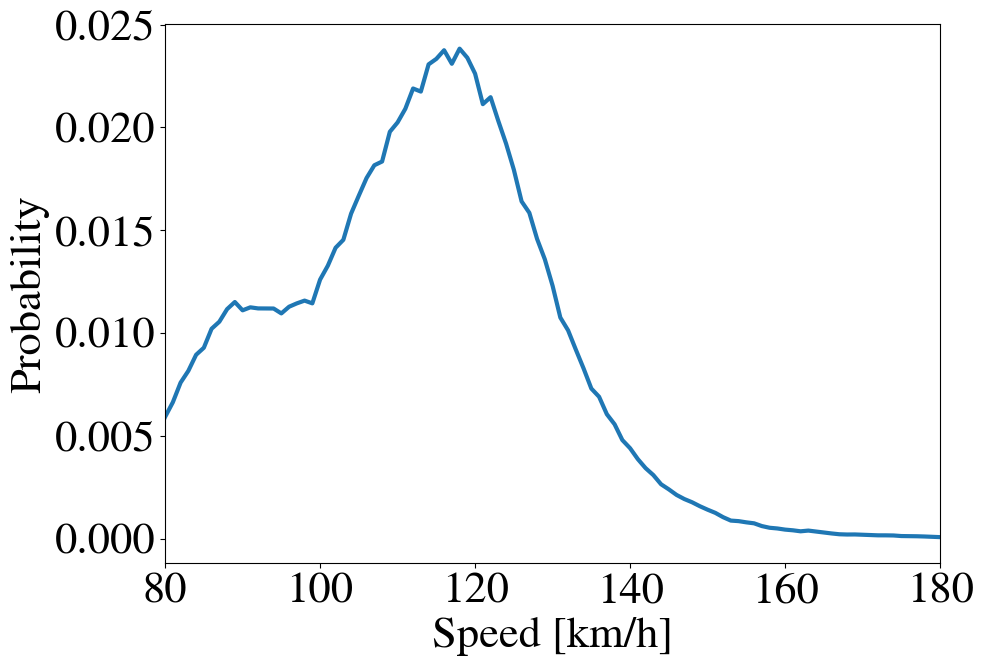}
  \caption{Speed distribution for cars}
\end{subfigure}
\hfill
\begin{subfigure}{.32\textwidth}
  \centering
  \includegraphics[width=\columnwidth]{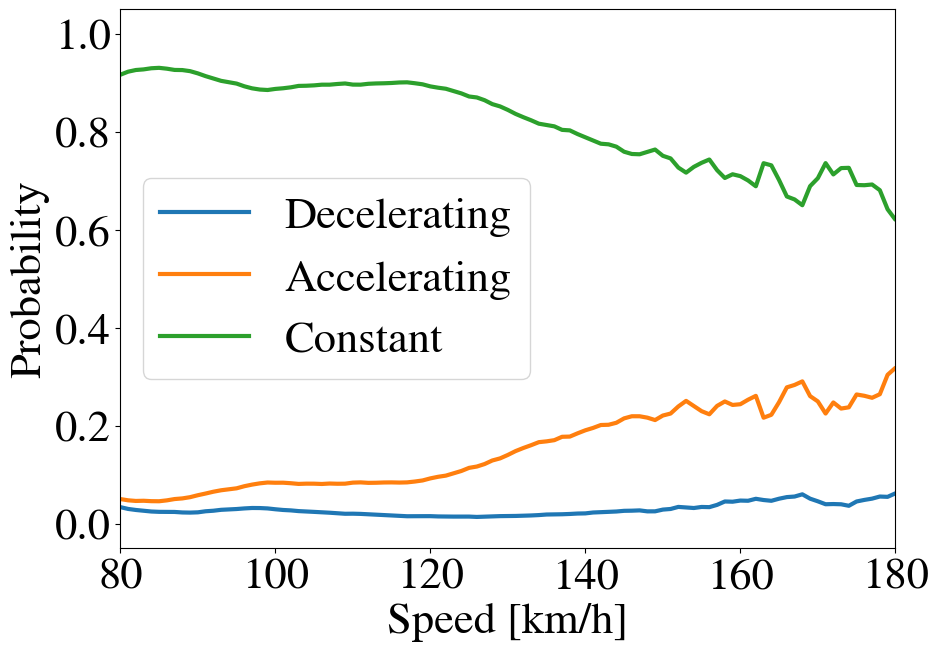}
  \caption{Distribution of different driving modes}
\end{subfigure}
\hfill
\begin{subfigure}{.32\textwidth}
  \centering
  \includegraphics[width=\columnwidth]{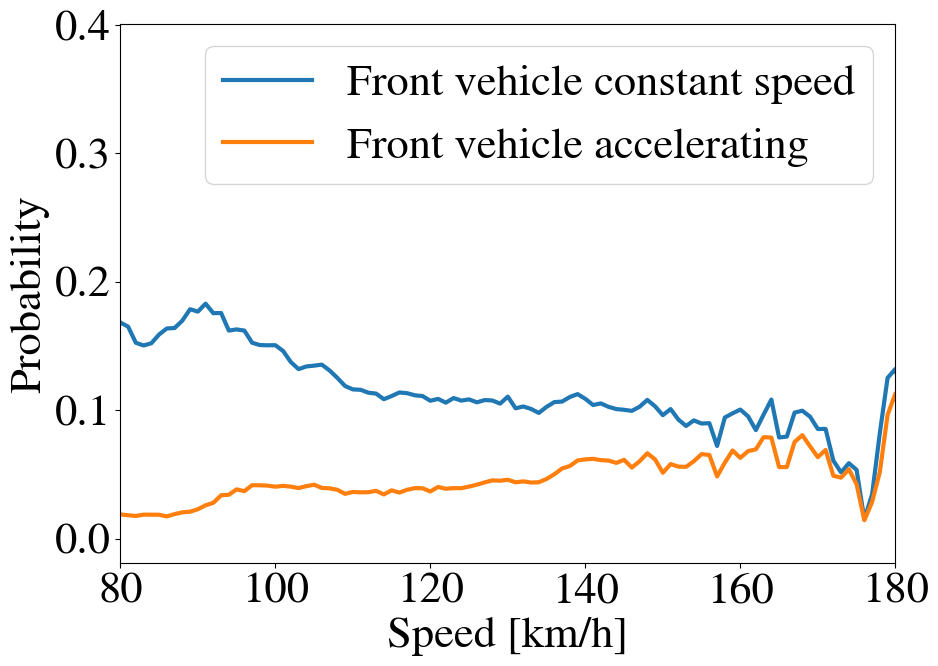}
  \caption{Distribution of two cars with TTC$\leq$5s}
\end{subfigure}
\caption{Naturalistic Driving Data from HighD dataset}
\label{fig:highddata}
\vspace{-0.4cm}
\end{figure*}

%Finally it is worth mentioning that the Lyft perception dataset does not contain velocity information for the other objects. As a result, we do not make use of the model refinements presented in Figure~\ref{fig:model_refinement}, instead the basic model shown in Figure~\ref{fig:model} is used.

In the following, we will discuss first some baseline numbers for human drivers and demonstrate how our model can be used to obtain perception quality requirements based on a target MTBF. Afterwards, we will show how a vehicle-level failure rate can be derived using perception data, and finally a discussion will conclude this section.

\vspace{-0.1cm}
\subsection{Human Driver Baseline}
For highway driving, various countries and official authorities gather data, which allows to obtain the failure rate or MTBF (time between two collisions) for human drivers. As often only severe events get reported, we focus in this work on collisions with S2 and S3 severity according to the ISO 26262, i.e. with severe or fatal injuries. According to an accident report of the German Federal Statistical Office~\cite{2020StatBuAmtUnfaelle}, 19.980 accidents (S2 or S3) were caused in 2019 on German highways, while 252.8 billion kilometers were driven on these roads. Assuming an average speed of 100\,km/h, which is reasonable according to~\cite{lohe2016geschwindigkeiten}, the MTBF of a human driver on a German highway is $1.3 \times 10^5$ hours. This is comparable to the numbers reported by the NHTSA (National Highway Traffic Safety Administration) for accidents on roads with speeds beyond 60\,mph in US~\cite{2019NHTSAAccidents}.

Based on these reports, it is also possible to obtain a collision severity estimation derived from the delta-velocity of the two road users at collision time. According to~\cite{JUREWICZ20164247} and~\cite{2019NHTSAAccidents}, a delta velocity of more than 30\,km/h results in a severity of S2 or S3\footnote{Note that this is only a first order estimation and ignores that often secondary collisions with even higher severity may occur}.

\subsection{Naturalistic Driving Data from HighD}
As illustrated in Figure~\ref{fig:model}, our model requires naturalistic driving information. Due to the restrictions in this study on highway driving, lane following situations and Type~II perception errors (perception misses), $m$ and $t$ can be ignored in Equation~\eqref{eq:extendedModel}. Thus, Equation~\eqref{eq:extendedModel} can be simplified to
\begin{equation}
\lambda = \sum_i p_i \times \lambda_i \times p_{S_i},
\end{equation}
where $p_i$ is the probability of driving with a speed in speed range $i$, $\lambda_i$ is the perception miss rate in speed range $i$ and $p_{S_i}$ is the probability of being in potentially dangerous traffic situations for this speed range. In this regard, the relevant situations are as depicted in Figure~\ref{fig:rds}, namely that the AV is within close distance to a leading vehicle, which either brakes, accelerates or drives with constant speed. In the latter two cases the AV has to be faster than the lead vehicle to be relevant. For this reason, we define ``close'' to the lead vehicle, as a TTC\footnote{In this work, we assume that the rear vehicle could accelerate with 2\,m/s² when calculating the time-to-collision} (time-to-collision) of less than 5\,seconds. Thus, $p_{S_i}$ can be obtained as follows:
\begin{equation}
p_{S_{i}} = p_{d_i} + p_{a_i} \times p_{a_{TTC,i}} + (1-p_{a_i}-p_{d_i}) \times p_{c_{TTC,i}},
\end{equation}
where $p_{d_i}$ is the probability of a lead car decelerating in speed range $i$, $p_{a_i} \times p_{a_{TTC,i}}$ is the probability that the lead car is sufficiently close ($p_{a_{TTC,i}}$) and accelerates ($p_{a_i}$), and $(1-p_{a_i}-p_{d_i}) \times p_{c_{TTC,i}}$ is the probability that the lead car is sufficiently close ($p_{c_{TTC,i}}$) and drives with constant speed ($p_{c_i} = 1-p_{a_i}-p_{d_i}$).

All of this information can be extracted from the HighD dataset, which is a drone-recorded dataset comprising 150 hours of recording with thousands of vehicles. A subset of the data, relevant for our model, is depicted in Figure~\ref{fig:highddata}. Figure~\ref{fig:highddata}(a) shows the speed distribution, (b) depicts the conditional probabilities for a vehicle accelerating ($p_a$), decelerating ($p_c$) or driving with a constant speed ($p_d$), and the conditional probability of following a vehicle with relevant distance is represented in Figure~\ref{fig:highddata}(c).

As can easily be seen, it is advisable to use multiple speed ranges, as for example the probability of a vehicle accelerating increases significantly beyond 130\,km/h, while the chance of a car following another car with less than 5 second TTC drops considerably for speeds faster than 100\,km/h. As a result, we use the following speed ranges for this evaluation: 80-100\,km/h, 100-130\,km/h and 130-180\,km/h. Situations with faster or slower velocities were discarded from the analysis. 

For theses speed ranges, the speed probability as well as the probabilities for the different potentially dangerous traffic situations are given in Table~\ref{tab:situationprobs}. As one can infer from this table, the probability of a potentially dangerous situation is between 10\,\% to 15\,\%, and the most likely speed range for the selected mission profile is the range between 100\,-\,130\,km/h with 64.0\,\%. In other words, in 64.0\,\% of the time on a highway a vehicle drives in this speed range, and finds itself in 2.1\,\% of this time in a situation behind a decelerating vehicle, in 0.3\,\% of the time behind a slower but accelerating vehicle, and in 15.2\,\% of the time the vehicle follows another slower vehicle that drives with constant speed. The remaining 82.4\,\% of the time in this speed range, the vehicle is not within a potentially dangerous traffic situation.

\begin{table}[b!]
\vspace{-0.25cm}
\centering
\begin{tabular}{|c||c|c|c|}
\hline
Speed [km/h] & 80 - 100 & 100 - 130 & 130 - 180 \\
\hline
\hline
Speed probability $\left[ p_i \right]$ & 0.234 & 0.640 & 0.126 \\
\hline
\hline
Lead vehicle decelerating $\left[ p_{d_i} \right]$ & 0.028 & 0.021 & 0.023 \\
\hline
Lead vehicle accelerating   & \multirow{ 2}{*}{0.001} & \multirow{ 2}{*}{0.003} & \multirow{ 2}{*}{0.004} \\
$\left[ p_{a_i} \times p_{a_{TTC,i}} \right]$ & & & \\
\hline
Lead vehicle constant speed & \multirow{ 2}{*}{0.279} & \multirow{ 2}{*}{0.152} & \multirow{ 2}{*}{0.088} \\
$\left[ p_{c_i} \times p_{c_{TTC,i}} \right]$ & & & \\
\hline
\hline
Total situation probability [$p_{S_i}$] & 0.308 & 0.176 & 0.115 \\
\hline
\end{tabular}
\caption{Probabilities of relevant driving situations and probability for different speed ranges}
\label{tab:situationprobs}
\end{table}

\begin{figure*}[t!]
\centering
\includegraphics[width=0.95\textwidth]{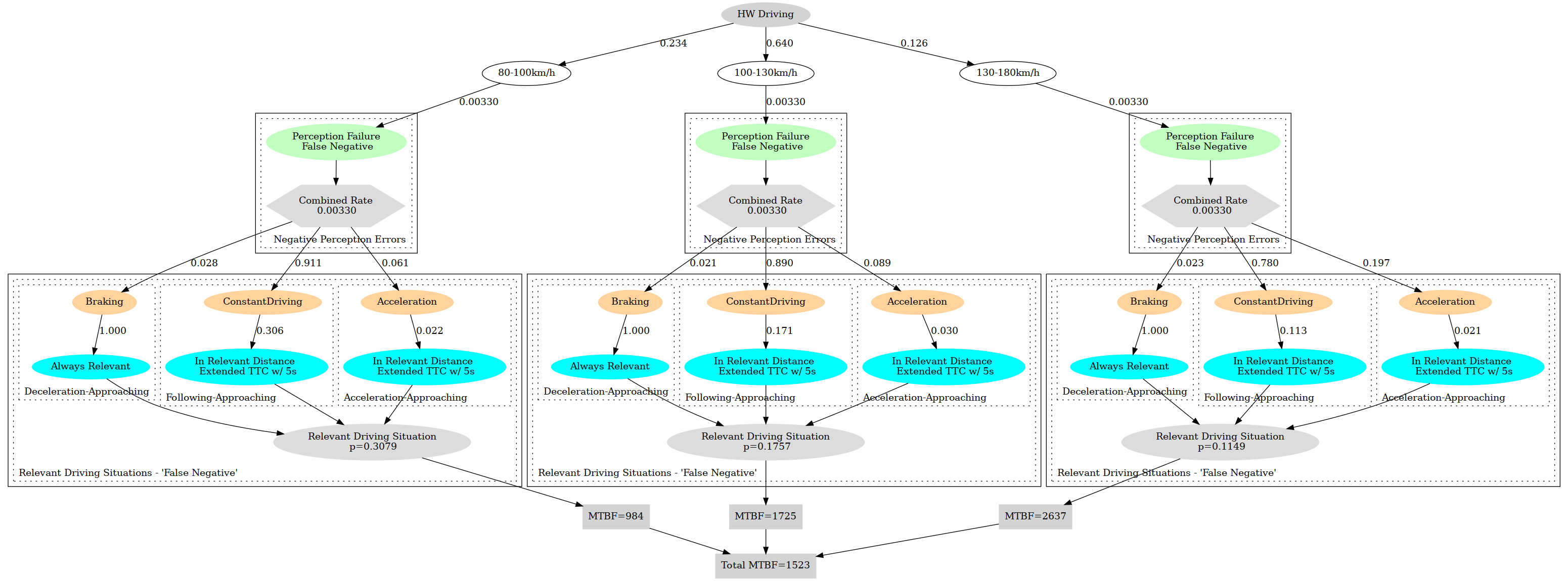}
\caption{MTBF model tree for highway driving based on driving data from HighD and perception data from Lyft}
\label{fig:results}
\vspace{-0.45cm}
\end{figure*}

\subsection{Extracting Perception Quality Requirements}
Given the situation probabilities derived in the previous subsection, our model illustrated in Figure~\ref{fig:model} can be used to derive the required perception quality (error rates) for a target failure rate at vehicle-level. Assuming that perception misses (False Negatives) are velocity independent, it follows with Equation~(9) that:
\begin{equation}
\lambda = \sum_i \lambda_i \times p_i \times p_{S_i} = \hat{\lambda} \times \sum_i p_i \times p_{S_i} =:  \hat{\lambda} \times \kappa
\end{equation}
where $ \hat{\lambda} = \lambda_i$ is the perception error rate, $p_i$ the probability of driving within speed range $i$, $p_{S_i}$ is the probability of having a potentially dangerous traffic situation in speed range $i$, and $\kappa$ is defined as $\sum p_i \times p_{S_i}$.

\begin{table}[b!]
\centering
\vspace{-0.2cm}
\begin{tabular}{|c|c|}
\hline
Target Vehicle-Level MTBF & Perception Error Rate \\
\hline
\hline
$10^4$ hours                & $5.0 \times 10^{-4}$ errors/hour\\
$10^5$ hours (= human MTBF) & $5.0 \times 10^{-5}$ errors/hour\\
$10^6$ hours                & $5.0 \times 10^{-6}$ errors/hour\\
$10^7$ hours                & $5.0 \times 10^{-7}$ errors/hour\\
\hline
\end{tabular}
\caption{Required perception quality (here: rate of perception misses) to achieve a target vehicle-level MTBF}
\label{tab:extrapolation}
\end{table}

Consequently, by defining a target value for $\lambda$, $\hat{\lambda}$ can be derived. For the speed ranges used in this study, and according to Table~\ref{tab:situationprobs}, $\kappa = 0.176$, which leads to the results presented in Table~\ref{tab:extrapolation}. In other words, only 19.9\,\% of all severe perception errors can manifest as vehicle-level failures. As a result, severe perception misses have to be less frequent than $10^{-4}$ to achieve an overall vehicle-level MTBF that is comparable to human drivers (MTBF = $1.3 \times 10^5$ for severe accidents as explained in Section~IV.A).

Please note that the starting assumption that perception misses are velocity-independent may not hold entirely in practice. On one hand, the physical properties of any sensing system do not change with speed, thus the probability for missing an object that is n meters away from the sensor does not change either. On the other hand, the further objects are away from the sensor, the more likely these will not be detected. At the same time, higher speeds require longer safety distances, and thus perception misses that are further away from the AV become relevant with increasing speeds. Thus, the assumptions holds as long as the sensor reach is more than the required safety distances (e.g. according to RSS), and as an AV should never drive with such speeds, the assumption is reasonable.

\subsection{From Perception Errors to Vehicle-Level Failures}
Our model can also be used to derive the vehicle-level failure rates from perception error rates, as explained in Section~\ref{sec:approach}. To illustrate this, we have used the Lyft dataset~\cite{LyftDS} to obtain perception error rates using a LiDAR based object detector, called PointPillars~\cite{PointPillars}. The situational probabilities are again extracted from the HighD dataset, as explained in the previous subsections. Please note that there is no publicly available large-enough perception dataset for highways available, thus in replacement we have chosen the Lyft dataset.

To estimate the MTBF for AV accidents with S2 or S3 severity, we considered only severe perception misses according to Definition 3. Furthermore, as the Lyft perception dataset does not provide detailed object velocity information, we assume a potential worst-case behavior, i.e. that the leading vehicle stands still and that the AV drives with maximum allowed speed for the given road type. Lyft covers 25200 Frames (equivalent to 1.4 hours of recording). We evaluated recorded LiDAR data using PointPillars and found 3 safety-relevant perception misses with severity S2 or S3, with a total duration of 17 frames. Thus, the severe perception miss rate (frames with misses per second) is $ 17/5040\,s= 12.1$\,errors/second. Assuming that this is velocity independent (as argued in the previous subsection), this can be combined with the situational probabilities from the HighD dataset. Consequently, as depicted in Figure~\ref{fig:results}, the overall vehicle-level MTBF is 1523\,second (0.4\,hours), which is significantly below the targeted MTFB of a human driver ($10^5$\,hours). In this regard, it is worth noting that the key contributor to this value seems to be the slowest speed range, due to a higher rate of potentially dangerous traffic situations.% vehicle-level failure rates are very similar for all velocity ranges, and that there is not one that can be claimed to be the key contributor.

Please note that the chosen perception system is an academic realization with a recall of around 90\,\%. In a real AV, a much more comprehensive perception solution could be used comprising of multiple camera, LiDAR and Radar sensors. Consequently, it is expected that in reality, the perception quality will be significantly better. The realization at hand was only chosen to illustrate how our model can be used and fed with data.

\subsection{Discussions}
In the preceding subsections, various results were shared, and different use cases of our proposed model were illustrated. To draw adequate conclusions, it is however important to keep several key aspects in mind, which we highlight next.

First, when using such a probabilistic model in combination with large datasets, it is important to ensure that the datasets contain sufficient and relevant information for the mission profile under evaluation to derive results with reasonable confidence values. For example, although the HighD dataset is very comprehensive, convergence studies show that the speed distribution (probability density function) changes considerably with every track file evaluated (e.g. using a Kolmogorov-Smirnov test). In other words, the speed distribution obtained from the HighD dataset is not a sufficient representation of naturalistic highway driving behavior in Germany. A similar study is required for the perception database, where it is advisable to evaluate if all relevant traffic constellations are represented~\cite{hauer2019did}, or that all illumination and weather conditions relevant for the mission profile are sufficiently covered.

If the available data is limited, it is advisable to reduce the model complexity, for example by using fewer, yet broader speed ranges, or to not differentiate between daylight driving and driving at night. In this case, reasonably chosen assumptions might be the better choice than using a fine-grained model that produces results with poor confidence. Of course, a coarser model will come at the cost of a less precise MTBF estimation (higher variance, but better confidence).

Another way of handling limited dataset is to apply data augmentation. For example, by using a worst-case assumption that all not-detected objects stand still, the analysis will provide a lower-bound for the vehicle-level MTBF values. Yet, it is important to note that performing data augmentation may bias the evaluation, resulting in an over-representation of worst-case situations. Hence, data augmentation has to be applied carefully, and should be considered for the perception error rate estimation as well as for the corresponding potentially dangerous driving situations. For instance, one can split the model into two sub-trees, one with augmentation (stand still) and add to the potentially dangerous traffic situation the probability that a vehicle approaches a standing obstacle, and another sub-tree without augmentation. 

It is also important to be aware that e.g the HighD naturalistic driving data is based on human drivers, and that an AV behave entirely different which may result in fewer/more potentially dangerous traffic situations. This difference has to be considered for the failure rate estimation.

Finally, it is worth discussing the consideration of an error duration. In this study, we handle every frame as it is an entirely new situation. For example, if in 1000 consecutive frames, a single perception miss event occurs that lasts for 500 frames, we consider this as 500 misses resulting in a perception error rate of 0.5. However, if there are 100 miss events each with a duration of 5 frames, the error rate is also 0.5. Yet, in practice, there is a significant difference among both cases. For this reason, we plan to extend our model in future work to also consider the duration of perception errors as well as potentially dangerous traffic situations.

%\vspace{-0.15cm}
\section{Conclusion}
\label{sec:conclusion}
AVs are on the edge to become reality. To reach the final step, it will be important for manufacturers to prove that their AVs achieve the required safety goals and among those ensure that the required vehicle-level failure rates (i.e. collisions) are below the requirements from governments and authorities. However, many different components can contribute to the overall vehicle-level failure rate raising the need for a comprehensive model that takes all of these into account without the need to perform exhaustive road testing for every licensing request. It is pragmatic that only such models are able to make the AV licensing process and the software update cycles scalable.

As perception errors will become the main error source in AVs assuming the planning components can be protected by safety models such as RSS~\cite{shalev2017formal}, it is important to estimate the impact of the perception errors on the overall vehicle-level failures (collisions). In this work, we presented such a formal model, which enables AV developers to derive a vehicle-level MBTF estimation given certain mission profiles, naturalistic driving data and perception error rates. In addition, it can be used to derive requirements for the perception quality given a target vehicle-level MTBF value. Both use cases were studied in an exemplary manner in this paper. % using publicly available datasets, which demonstrated the feasibility and flexibility of the proposed model. 
For example, using publicly available driving data (HighD) we obtained that the perception error rate should not exceed $10^{-5}$ errors/hour to achieve an MTBF above human performance. The results demonstrated the feasibility and flexibility of the proposed model, which can be applied to any AV perception system. % on selected mission profiles to determine vehicle-level MTBF.

\vspace{-0.1cm}

\bibliographystyle{IEEEtran}
\bibliography{main}

\begin{thebibliography}{10}
\providecommand{\url}[1]{#1}
\csname url@samestyle\endcsname
\providecommand{\newblock}{\relax}
\providecommand{\bibinfo}[2]{#2}
\providecommand{\BIBentrySTDinterwordspacing}{\spaceskip=0pt\relax}
\providecommand{\BIBentryALTinterwordstretchfactor}{4}
\providecommand{\BIBentryALTinterwordspacing}{\spaceskip=\fontdimen2\font plus
\BIBentryALTinterwordstretchfactor\fontdimen3\font minus
  \fontdimen4\font\relax}
\providecommand{\BIBforeignlanguage}[2]{{%
\expandafter\ifx\csname l@#1\endcsname\relax
\typeout{** WARNING: IEEEtran.bst: No hyphenation pattern has been}%
\typeout{** loaded for the language `#1'. Using the pattern for}%
\typeout{** the default language instead.}%
\else
\language=\csname l@#1\endcsname
\fi
#2}}
\providecommand{\BIBdecl}{\relax}
\BIBdecl

\bibitem{shalev2018vision}
S.~Shalev-Shwartz, S.~Shammah, and A.~Shashua, ``Vision zero: can roadway
  accidents be eliminated without compromising traffic throughput,'' 2018.

\bibitem{robotaxi2022}
Mobileye, ``{Launching our Self-Driving Robotaxi in Germany},'' 2021.

\bibitem{SaFAD2019}
Aptiv, Audi \emph{et~al.}, ``{Safety First for Automated Driving},'' Tech.
  Rep., 2019.

\bibitem{di2017ethics}
U.~Di~Fabio, M.~Broy \emph{et~al.}, ``Ethics commission automated and connected
  driving,'' \emph{Federal Ministry of Transport and Digital Infrastructure of
  the Federal Republic of Germany}, 2017.

\bibitem{weastMTBF}
\BIBentryALTinterwordspacing
J.~Weast, ``{Sensors, Safety Models and A System-Level Approach to Safe and
  Scalable Automated Vehicles},'' \emph{CoRR}, vol. abs/2009.03301, 2020.
  [Online]. Available: \url{https://arxiv.org/abs/2009.03301}
\BIBentrySTDinterwordspacing

\bibitem{shalev2017formal}
S.~Shalev-Shwartz, S.~Shammah, and A.~Shashua, ``On a formal model of safe and
  scalable self-driving cars,'' \emph{arXiv:1708.06374}, 2017.

\bibitem{ieeeAVSafety}
``{IEEE P2846: A Formal Model for Safety Considerations in Automated Vehicle
  Decision Making},'' 2022.

\bibitem{buerkle2021safe}
C.~Buerkle, F.~Oboril \emph{et~al.}, ``{Safe Perception: On Relevance of
  Objects for Vehicle Safety},'' in \emph{2021 IEEE International Intelligent
  Transportation Systems Conference (ITSC)}.\hskip 1em plus 0.5em minus
  0.4em\relax IEEE, 2021.

\bibitem{berk2019safety}
M.~Berk, ``{Safety Assessment of Environment Perception in Automated Driving
  Vehicles},'' Ph.D. dissertation, Technische Universit{\"a}t M{\"u}nchen,
  2019.

\bibitem{krajewski2018highd}
R.~Krajewski, J.~Bock \emph{et~al.}, ``The highd dataset: A drone dataset of
  naturalistic vehicle trajectories on german highways for validation of highly
  automated driving systems,'' in \emph{International Conference on Intelligent
  Transportation Systems (ITSC)}.\hskip 1em plus 0.5em minus 0.4em\relax IEEE,
  2018.

\bibitem{li2015driving}
G.~Li, S.~E. Li \emph{et~al.}, ``Driving maneuvers analysis using naturalistic
  highway driving data,'' in \emph{2015 IEEE 18th International Conference on
  Intelligent Transportation Systems}.\hskip 1em plus 0.5em minus 0.4em\relax
  IEEE, 2015, pp. 1761--1766.

\bibitem{2020StatBuAmtUnfaelle}
S.~B. (Destatis), ``Verkehrsunfälle,'' \emph{Verkehr}, 2020.

\bibitem{lohe2016geschwindigkeiten}
U.~L{\"o}he, ``Geschwindigkeiten auf bundesautobahnen in den jahren 2010 bis
  2014,'' \emph{Schlussbericht Zum AP-Projekt F}, vol. 1100, 2016.

\bibitem{2019NHTSAAccidents}
``{National Highway Traffic Safety Administration - Traffic Safety Facts
  2019},'' 2019.

\bibitem{JUREWICZ20164247}
C.~Jurewicz, A.~Sobhani \emph{et~al.}, ``Exploration of vehicle impact speed
  – injury severity relationships for application in safer road design,''
  \emph{Transportation Research Procedia}, vol.~14, pp. 4247--4256, 2016,
  transport Research Arena TRA2016.

\bibitem{LyftDS}
R.~Kesten, M.~Usman \emph{et~al.}, ``Level 5 perception dataset 2020,''
  \url{https://level-5.global/level5/data/}, 2019.

\bibitem{PointPillars}
A.~H. {Lang}, S.~{Vora} \emph{et~al.}, ``Pointpillars: Fast encoders for object
  detection from point clouds,'' in \emph{2019 IEEE/CVF Conference on Computer
  Vision and Pattern Recognition (CVPR)}, 2019.

\bibitem{hauer2019did}
F.~Hauer, T.~Schmidt \emph{et~al.}, ``Did we test all scenarios for automated
  and autonomous driving systems?'' in \emph{2019 IEEE Intelligent
  Transportation Systems Conference (ITSC)}.\hskip 1em plus 0.5em minus
  0.4em\relax IEEE, 2019, pp. 2950--2955.

\end{thebibliography}

\end{document}